\documentclass[letterpaper, 10 pt, conference]{ieeeconf}  

\IEEEoverridecommandlockouts                              

\usepackage{times}
\usepackage{epsfig}
\usepackage{graphicx}
\usepackage{amsmath}
\usepackage{hyperref}

\usepackage{subcaption}
\DeclareCaptionSubType*[Alph]{table}
\DeclareCaptionLabelFormat{mystyle}{Table~\bothIfFirst{#1}{ ̃}#2}
\usepackage{multirow}
\usepackage{array}
\newcolumntype{P}[1]{>{\centering\arraybackslash}p{#1}}

\usepackage{xcolor}
\usepackage{comment}

\title{\LARGE \bf
F-SIOL-310: A Robotic Dataset and Benchmark for Few-Shot Incremental Object Learning
}

\author{Ali Ayub$^{1}$ and Alan R. Wagner$^{2}$
\thanks{$^{1}$Department of Electrical Engineering,
        The Pennsylvania State University, State College, PA 16802, USA
        {\tt\small aja5755@psu.edu}}%
\thanks{$^{2}$Department of Aerospace Engineering, The Pennsylvania State University,
        State College, PA 16802, USA
        {\tt\small alan.r.wagner@psu.edu}}%
}

\begin{document}

\maketitle
\thispagestyle{empty}
\pagestyle{empty}

\begin{abstract}
\label{sec:Abstract}
Deep learning has achieved remarkable success in object recognition tasks through the availability of large scale datasets like ImageNet. However, deep learning systems suffer from catastrophic forgetting when learning incrementally without replaying old data. For real-world applications, robots also need to incrementally learn new objects. Further, since robots have limited human assistance available, they must learn from only a few examples. However, very few object recognition datasets and benchmarks exist to test incremental learning capability for robotic vision. Further, there is no dataset or benchmark specifically designed for incremental object learning from a few examples. To fill this gap, we present a new dataset termed F-SIOL-310 (Few-Shot Incremental Object Learning) which is specifically captured for testing few-shot incremental object learning capability for robotic vision. We also provide benchmarks and evaluations of 8 incremental learning algorithms on F-SIOL-310 for future comparisons. Our results demonstrate that the few-shot incremental object learning problem for robotic vision is far from being solved.


\end{abstract}
\section{INTRODUCTION}
\label{sec:introduction}
\noindent Humans have the ability to learn new concepts continually over their lifetime from only a few examples. With robots increasingly becoming an integral part of the society for a variety of different roles, such as household robots \cite{Matari17}, they should also learn about new concepts continually to adapt to their dynamic environments. Further, since robots have limited human assistance in real-world environments, they must learn from only a few examples \cite{lesort20}. 
The ultimate goal of this paper is to develop an object recognition dataset for continual learning with a few examples on robots. 

One of the reasons for deep learning's remarkable success in object recognition is the availability of large-scale object datasets like ImageNet \cite{Russakovsky15} and CIFAR-100 \cite{Krizhevsky09}. These datasets, however, have been designed for batch learning in which a model is trained on the data of all the classes in one batch and then evaluated on a separate test set. Further, these datasets generally contain objects in ideal conditions, such as no blurriness or transparency etc. In contrast, robots generally do not have access to a large set of idealistic, labeled images in the real world. Robots mostly have to continually acquire data through their own camera autonomously in different conditions which often leads to less than perfect images. Further, to get labels for the captured object images, robots have to ask for human assistance. Human teachers are usually unwilling to answer enormous numbers of questions, hence robots may only have a few labeled object images to learn from. Thus, large scale datasets like ImageNet and CIFAR-100 are not suitable for evaluating incremental learning approaches for robot vision.

Most state-of-the-art (SOTA) incremental learning approaches \cite{kirkpatrick17,Ayub_ICML_20,Rebuffi_2017_CVPR,Li18}, however, have been tested on less complex datasets like MNIST \cite{Lechun98} or CIFAR-100 \cite{Krizhevsky09}. Further, most current incremental learning approaches \cite{kirkpatrick17,Ayub_ICML_20,Rebuffi_2017_CVPR,Li18} are not designed for few-shot incremental learning (FSIL), which is a challenging, but realistic, learning problem for robotics applications that was first considered in \cite{Ayub_2020_CVPR_Workshops}. FSIL is necessary for real-world robotics applications like household cleaning robots or industrial packaging robots or anywhere a robot may have to quickly learn new objects from a few examples provided by humans. For FSIL, an incremental learning model is required to learn object classes incrementally but with only a few (usually 5 or 10) training examples per class. CBCL \cite{Ayub_2020_CVPR_Workshops} was specifically designed for FSIL, however, it was primary tested on the CIFAR-100 dataset since no dataset had been specifically created for FSIL. 


\begin{figure}[t]
\centering
\includegraphics[scale=0.05]{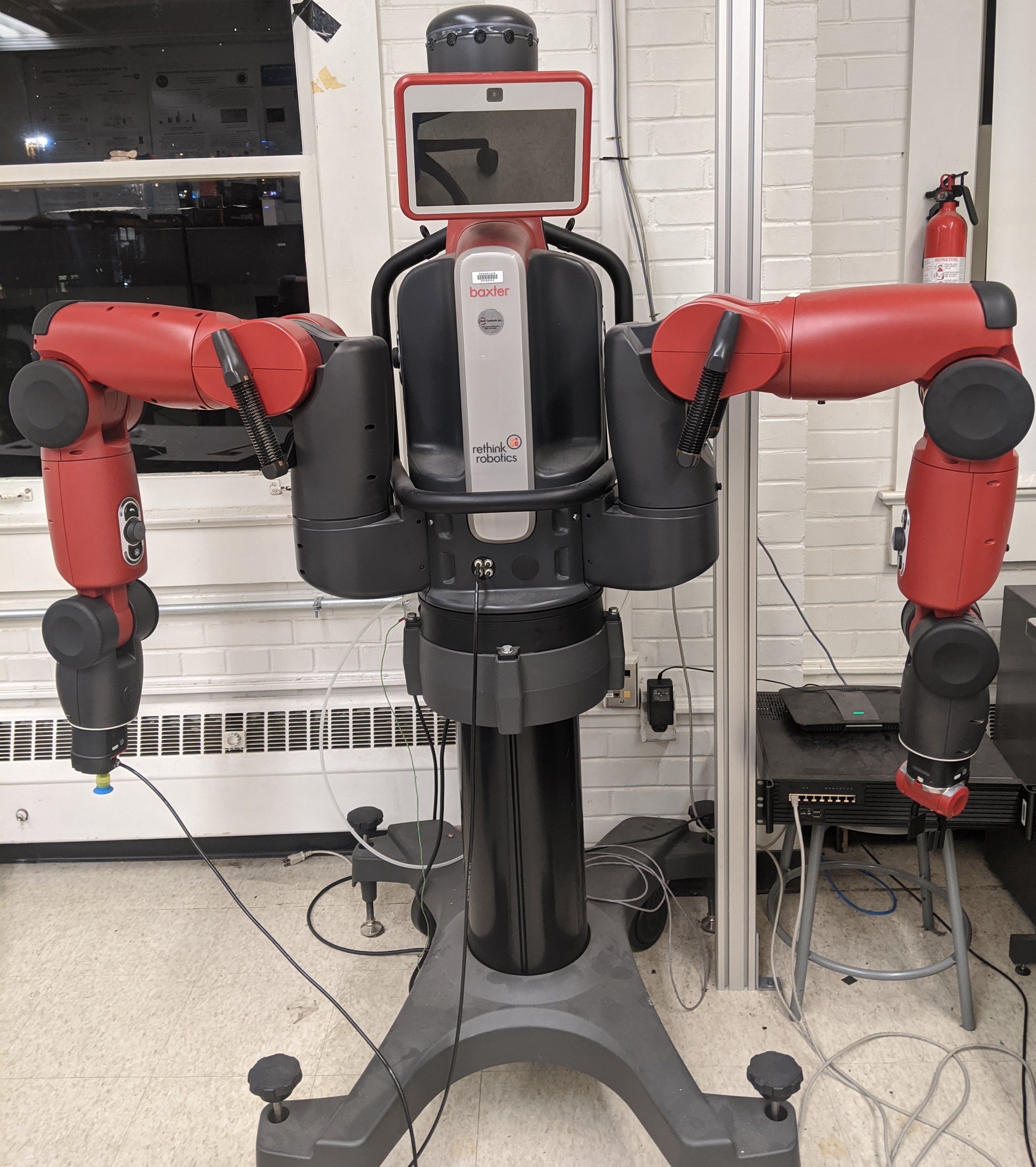}
\caption{\small Baxter robot used to capture the F-SIOL-310 object dataset.}
\label{fig:baxter}
\end{figure}

\begin{table*}
\small
\begin{tabular}{P{3.0cm}P{1.0cm}P{1.0cm}P{1.8cm}P{2.0cm}P{1.8cm}P{1.3cm}P{1.8cm}}
     \hline
     \textbf{Datasets} & \textbf{Classes} & \textbf{Objects} & \textbf{Acquisition} & \textbf{Train/Test Difference} & \textbf{FSIL Benchmarks} & \textbf{Scale Variation} & \textbf{Transparency}\\
     \hline
     COIL-100 \cite{nene96} & - & 100 & turntable & No & No & No & No\\
     NORB \cite{lechun04} & 5 & 25 & turntable & No & No & No & No\\
     BigBIRD \cite{singh14} & - & 100 & turntable & No & No & No & No\\
     ROD \cite{lai11} & 51 & 300 & turntable & No & No & No & No\\
     CORe-50 \cite{lomonaco17} & 10 & 50 & hand held & No & No & No & No\\
     ARID \cite{reza18} & 51 & 153 & robot & No & No & Yes & No\\
     OpenLORIS \cite{she19} & 19 & 69 & robot & No & No & Yes & No\\
     \hline
     \textbf{F-SIOL-310} & \textbf{22} & \textbf{310} & \textbf{robot} & \textbf{Yes} & \textbf{Yes} & \textbf{Yes} & \textbf{Yes}\\
 \hline
 \end{tabular}
 \caption{Comparison of F-SIOL-310 with other robotics object recognition datasets.}
 \label{tab:datasets}
 \end{table*}

In order to close the gap between existing robot vision research and real-world applications, we propose a new RGB dataset termed F-SIOL-310. A Baxter robot (manufactured by Rethink robotics, see Figure \ref{fig:baxter}) was used to actively capture household objects on a table. The dataset is specifically designed for FSIL with only a small set of training images and a larger set of test images per object category captured by the robot using its own camera and it considers various other robot vision challenges as well, such as different object sizes, object transparency and a clear distinction between objects in the train and test sets. 
We provide extensive evaluations of the state-of-the-art incremental learning approaches using different backbone neural network architectures pre-trained on different datastets, on F-SIOL-310 for FSIL. 
The complete dataset is available at \url{https://tinyurl.com/yb38syd5}. This paper contributes:
\begin{enumerate}
    \item A new RGB object dataset termed F-SIOL-310. The dataset is collected using a diverse set of 310 household objects of various sizes. To the best of our knowledge, no other dataset has been developed for the FSIL problem which contains different objects belonging to a category in the train and test sets.   
    \item Benchmarks for evaluating FSIL with 5-shot and 10-shot settings using different test sets.
    \item An extensive set of evaluations of current SOTA incremental learning approaches on F-SIOL-310 with 5-shot and 10-shot incremental learning which show that the current approaches are far from being accurate enough for application in real world robotics problems.
\end{enumerate}

The remainder of the paper is organized as follows: Section \ref{sec:related_work} reviews the related work including incremental learning approaches and other robotics datasets for object recognition. Section \ref{sec:methodology} describes F-SIOL-310 in detail. Section \ref{sec:experiments} presents benchmarks and empirical evaluations of SOTA approaches on F-SIOL-310 for FSIL. Finally, Section \ref{sec:conclusion} offers conclusions and directions for future research.
\section{RELATED WORK}
\label{sec:related_work}

\subsection{Incremental Learning Techniques}
As discussed in \cite{she19}, a true incremental learning system should have the following characteristics: 1) it can learn new knowledge and patterns from new data; 2) it can avoid catastrophic forgetting and remember past knowledge; 3) it can generalize in response to new incoming data; 4) it should have few-shot learning capability so that it can learn from limited data; 5) it can learn from an infinite stream of data while keeping the memory footprint and learning time from growing drastically. She et al. \cite{she19} examined the first three capabilities in their work. In contrast, this paper evaluates current incremental learning techniques on all five characteristics.  
These characteristics require incremental learning algorithms to learn new object classes from limited data and to perform competitively on new and old object classes while keeping the system memory from growing substantially. 

\begin{figure*}
\centering
\includegraphics[scale = 0.3]{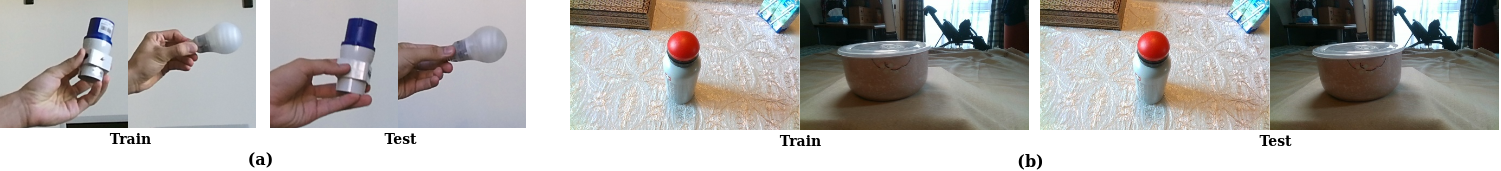}
\caption{\small Examples of training and test images in CORe-50 (a) and OpenLORIS (b) datasets. Note that same objects are used in train and test sets with similar backgrounds.}
\label{fig:openloris}
\end{figure*}

Creating highly accurate incremental object learning classifiers, however, is made more difficult by the catastrophic forgetting problem: the model (often a neural network) forgets the previously learned classes when learning new classes resulting in a significant decrease in classification accuracy. Most existing class-incremental learning methods avoid this problem by storing a portion of the training samples from previous classes and retraining the model (typically a neural network) on a mixture of the stored data and new data~\cite{Rebuffi_2017_CVPR,Castro_2018_ECCV}. However, storing real samples quickly exhausts memory capacity and limits performance for real-world applications. 
To avoid this problem, some incremental learning approaches use regularization techniques \cite{Li18,kirkpatrick17}. Although these approaches solve the memory storage issues, their performance is inferior to approaches that store old class data. Another genre of incremental learning approaches use generative memory such as autoencoders or GANs (Generative Adversarial Networks) \cite{Ostapenko_2019_CVPR,Ayub_ICML_20,Ayub_iclr20}. These approaches are currently only applicable on simpler datasets with a smaller number of classes. One of the main issues with all of these approaches is that they require a large number of labeled training images per class and cannot learn from few examples (FSIL problem). To the best of our knowledge, CBCL \cite{Ayub_2020_CVPR_Workshops,Ayub_IROS_20,Ayub_2020_BMVC,mokhtari20} is currently the only approach that tackles the FSIL problem. Although CBCL generates reasonable accuracy on FSIL problems, its accuracy depends on having a good, task-specific feature extractor trained on a large dataset.

One of the main problems with the incremental learning approaches mentioned above, is that they are evaluated on less complex datasets such as MNIST \cite{Lechun98} and CIFAR-100 \cite{Krizhevsky09}. These datasets contain idealistic images in constrained environments. For real-world applications, however, robots do not have access to images of objects in perfect conditions. Further, all of these datasets contain a large number of training images per class which is unrealistic for robots operating in real-world environments. Robots only have access to a limited number of training images that are labeled by a human. Hence, to develop more advanced incremental learning algorithms, there is a need for a dataset that includes realistic images taken by an embodied, situated robot. This dataset will act as a benchmark for future incremental object learning approaches tested on a robot.

\subsection{Related Datasets}
In recent years many robotics object recognition datasets have been publicly released for research. Table \ref{tab:datasets} shows a comparison of robotics object recognition datasets. Most of these datasets (COIL-100 \cite{nene96}, NORB \cite{lechun04}, BigBIRD \cite{singh14}, ROD \cite{lai11} 
and ARID \cite{reza18}) were designed for batch learning scenario, however they could potentially be used for incremental learning. To the best of our knowledge, only CORe-50 \cite{lomonaco17} and OpenLORIS \cite{she19} were specifically designed for incremental learning and released incremental learning benchmarks for robotics applications. 

The datasets COIL-100, NORB, BigBIRD and ROD present images of objects captured on a turntable in a systematically controlled environment with perfect lighting conditions and object views. Although these datasets are created for robotics applications, they miss many crucial challenges that robots face in real-world scenarios such as scale variation of objects and the presence of transparent objects. Further, NORB only has 5 object classes and 25 total objects, while COIL-100 and BigBIRD do not provide object classes and contain only 100 objects in total, which is extremely small compared to other datasets like CIFAR-100 or OpenLORIS. Classification results on such a small number of classes also do not suffer from catastrophic forgetting as shown in \cite{Rebuffi_2017_CVPR,Ayub_2020_CVPR_Workshops}. ROD is the only dataset that has more than 51 object classes and 300 total number of objects. However, each class contains only $\sim$6 objects. Another major issue with all of these datasets is that they contain the same objects in the train and test sets. Hence, algorithms trained on these datasets may overfit on the train set and nevertheless produce good results on the test set instead of learning generalized representations of objects. For real-world object recognition, algorithms need to learn general representations of objects so that they can recognize unseen objects of the same class \cite{Russakovsky15,pinto08}. Finally, since theses datasets have a small number of objects per class, there are too few different objects per class to create training and test sets that do not contain the same objects for few-shot incremental learning.



ARID is a recently captured robotics object recognition dataset which was captured by manually driving a robot through different environments. The dataset attends to many challenges of robot object recognition including that the images were captured by cameras on a robot and that the images include objects at different scales under different lighting conditions. However, this dataset was not designed for incremental object learning and it did not release any incremental learning benchmarks. ARID contains a large number of object classes (51), however the total number of objects per class are extremely small (3). Since there are only 3 objects per class, this dataset has to use same objects in the training and test sets and it cannot be used for learning from a few examples. Hence, this dataset cannot be used to evaluate Few-shot incremental learning algorithms.   

\begin{figure*}
\centering
\includegraphics[scale=0.375]{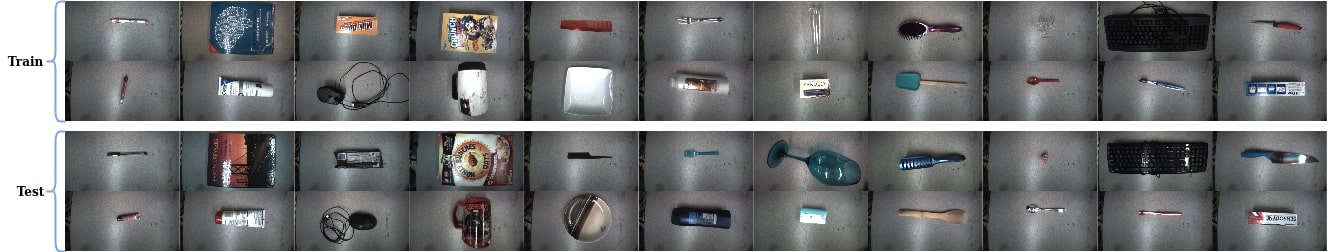}
\caption{\small Examples images of 22 object categories in F-SIOL-310. One object per category is shown from the training and test set splits for 5-shot incremental learning. Note the difference between the objects between the training and test sets.}
\label{fig:F_SIOL}
\end{figure*}

Core-50 is the first robotics dataset specifically designed for incremental object learning. It contains 11 video clips of objects in different backgrounds. CORe-50 was captured with the operator holding the objects in his hand, which makes it unsuitable for real-world autonomous robotics situations. In real-world scenarios, robots have to capture data through their own camera instead of asking for human assistance with every object. This dataset also does not consider many other challenges faced by robots in real-world scenarios such as scale variation of objects and the inclusion of transparent objects. Further, this dataset only consists of 10 object classes with 5 objects per class (total 50 objects). OpenLORIS is the latest robotics dataset that was designed for incremental object learning. Similar to ARID, this dataset was manually captured using cameras on an autonomous robot. OpenLORIS also captures the different challenges of robotics vision such as scale variation, lighting conditions etc and it provides benchmarks for incremental learning under different environmental conditions. Unfortunately, this dataset contains only 19 object classes with only $\sim$3 objects per class. The primary problems with CORe-50 and OpenLORIS are: 1) they contain same objects in the train and test sets. Although the objects are captured in slightly different environments, the objects themselves are the same in train and test sets (see Figure \ref{fig:openloris} for examples). 2) These datasets test the incremental learning capability of algorithms by providing the same objects in different increments captured in different backgrounds. This is contrary to the spirit of incremental learning where the algorithm does not have access to past objects when learning new objects. 3) these datasets are not suitable for testing Few-shot incremental learning and do not provide any FSIL benchmarks.

In contrast to these datasets, F-SIOL-310 contains different objects in the training and test sets (see Figure \ref{fig:F_SIOL}) and it is specifically designed for testing FSIL on robots. F-SIOL-310 contains 310 objects which is larger than all of the above mentioned datasets. Similar to ARID and OpenLORIS, F-SIOL-310 was captured using a camera on a robot. However, unlike ARID and OpenLORIS, F-SIOL-310 is captured using the arm camera of a manipulator robot (Baxter). Further, note that unlike CORe-50 and OpenLORIS, F-SIOL-310 only provides images of objects and not a set of videos with different views of objects. The reason is that in real-world scenarios it is not always easy for robots to get different views of objects and they usually have only one view of the object available. Both CORe-50 and OpenLORIS contain manually taken videos of objects from different views, which may not be possible for robots in real-world situations. 
\section{F-SIOL-310}
\label{sec:methodology}

\noindent F-SIOL-310 is a dataset specifically designed for few-shot incremental object learning on an autonomous robot. F-SIOL-310 consists of common household objects belonging to 22 classes: ballpoint, book, candy, cereal, comb, fork, glass, hair\_brush, hair\_clip, keyboard, knife, lipstick, lotion, mouse, mug, plate, shampoo, soap, spatula, spoon, toothbrush, toothpaste (see Figure \ref{fig:F_SIOL}). For each object class, we purchased 11-17 different objects leading to a total of 310 objects. 

The dataset can be used to perform class level or object level classification. Class level classification for FSIL is a much harder task than object-level classification, since the model is only given a few training samples per class but it is tested on a larger set which contains unseen objects. Hence, the model is required to learn a general representation of each object class from only a few training examples while learning classes incrementally, making the task challenging. In this paper, we focus on the class-level classification for FSIL experiments (Section \ref{sec:experiments}).

The images that compose the dataset were collected on a table-top. The objects were presented on the table in front of a Baxter robot and the robot's hand camera was used to capture the data. The robot's hand hovered over the table to caputre the objects. For each object two images from different views were captured, which resulted in a total of 22-34 RGB images per class. The total raw data consists of 620 RGB images, each of size 1020$\times$534. 

\begin{figure}[t]
\centering
\includegraphics[scale=0.45]{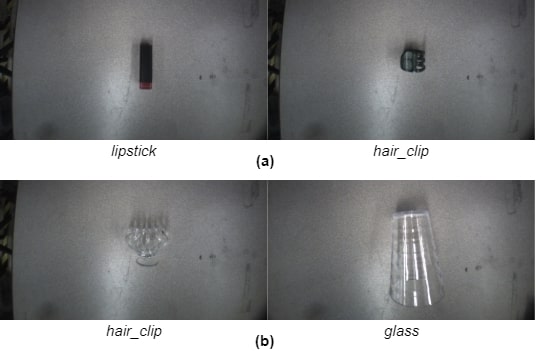}
\caption{\small Examples of images showing challenges in realistic robotic vision. (a) Objects belonging to categories \textit{lipstick} and \textit{hair\_clip} are relatively small. (b) Objects belonging to categories \textit{hair\_clip} and \textit{glass} are transparent.}
\label{fig:realistic_images}
\end{figure}

The images captured by the robot are realistic and not idealized. The background of the objects is not perfect since the table has several discolorations and some of the object sizes are rather small (Figure \ref{fig:realistic_images} (a)). Moreover, the lighting conditions are not ideal and some objects are transparent (Figure \ref{fig:realistic_images} (b)). Further, different objects are of different sizes or at different distance from the camera (Figure \ref{fig:F_SIOL}). Hence, it was not possible to crop the images to a fixed smaller size and we only provide the raw data and used it in our FSIL experiments (Section \ref{sec:experiments}). In case cropped objects are needed, we provide the bounding boxes for objects. The bounding boxes were generated by passing the images through the RetinaNet \cite{Lin_2017_ICCV}.

For evaluation on FSIL, we split the dataset into two different training and test sets for 5-shot and 10-shot incremental learning experiments. For 5-shot incremental learning, only 5 images per class are in the training set and the rest of the images are in the test set. Similarly, 10 images per class are in the training set for 10-shot incremental learning setting and the rest of the images are in the test set. Objects in the training set do not overlap with the test set i.e. training and test sets have different objects of the same class (see Figure \ref{fig:F_SIOL}). This setup is in accordance with realistic robotics scenarios, where the robot must learn the general concept of an object class and recognize unseen objects rather than just memorizing and recognizing the same objects.


\section{EXPERIMENTS}
\label{sec:experiments}


\noindent We evaluate various state-of-the-art (SOTA) incremental learning approaches on F-SIOL-310 for 5-shhot and 10-shot incremental learning scenarios. We further evaluate the accuracy of a variety of deep convolutional networks pre-trained on different datasets as backbone architectures in incremental learning approaches for FSIL experiments. 

\begin{figure*}
\centering
\includegraphics[scale=0.23]{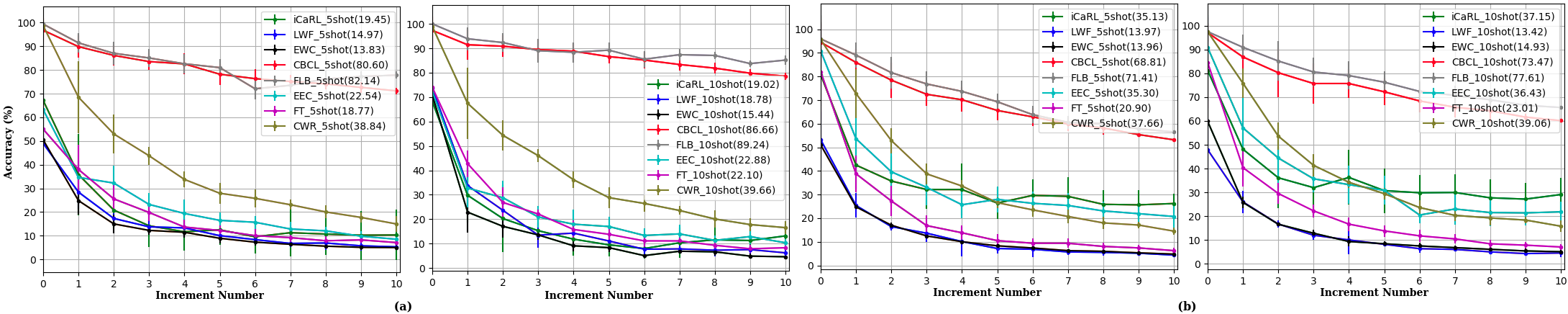}
\caption{\small Evaluation of 8 incremental learning approaches on F-SIOL-310 with ImageNet features (a) and CIFAR-100 features (b), for 5-shot (left) and 10-shot (right) incremental learning with 2 classes per increment in terms of classification accuracy (\%). Mean and standard deviations of the classification accuracy in each increment are reported across 10 runs with a different seed in each run. Average incremental accuracy is reported in parenthesis for each approach. FLB (in grey) is the batch learning upperbound that trains the model on all the previous and new class data. (Best viewed in color)}
\label{fig:few_shot_results}
\end{figure*}

\subsection{SOTA Incremental Learning Approaches}
\noindent We evaluate 8 incremental learning approaches (FLB \cite{Chen19}, FT, iCaRL \cite{Rebuffi_2017_CVPR}, LWF \cite{Li18}, EWC \cite{kirkpatrick17}, CWR \cite{lomonaco17}, CBCL \cite{Ayub_2020_CVPR_Workshops}, EEC \cite{Ayub_ICML_20}) on F-SIOL-310 for 5-shot and 10-shot incremental learning. Few-Shot learning baseline (FLB) uses the features from a pre-trained neural network and then trains a linear layer using the cross-entropy loss \cite{Chen19}. FLB is not designed for incremental learning, hence we use the complete training set of the previous classes and the new classes in each increment for training FLB. Thus, FLB is an upper bound depicting the performance of batch learning strategy for FSIL experiments. Fine-tuning (FT) is a naive approach in which the model is trained only on the data of the new classes in each increment. Other incremental learning approaches have been introduced in Section \ref{sec:related_work}. 

\subsection{Implementation Details}
The Pytorch deep learning framework \cite{torch19} and an Nvidia Titan RTX GPU were used for implementation and training of all neural network models. All of the input images were resized to $256 \times 256$ and randomly cropped to $224\times 224$ as the input to the network during training. ResNet18 \cite{He_2016_CVPR} pre-trained on ImageNet dataset \cite{Russakovsky15} was used as the backbone architecture for all the approaches. For all the models except CBCL, we trained the network with cross-entropy loss optimized with stochastic gradient descent (SGD) with momentum of 0.9 and a fixed learning rate of 0.1 for 40 epochs in each increment. For CBCL, we used a fixed distance threshold $D$=17 for \textit{Agg-Var} clustering and number of closest centroids for classification $k$=1 in each increment. 

For iCaRL, we allow it to store images of all the previous classes for both 5-shot and 10-shot incremental learning experiments which is similar to batch learning. CBCL and EEC are allowed to store a maximum of 44 centroids, making their memory consumption minimal as desired by incremental learning approaches. EWC, LWF and CWR do not require storage of images or feature vectors of previous classes.   
For both 5-shot and 10-shot incremental learning experiments, we divided the training set into 11 batches with 2 classes per batch learned in each increment. 
For evaluation, we tested each approach in each increment on all the classes it had learned so far, resulting in decreasing accuracy curves. We also report \textit{average incremental accuracy} which is the average of all the accuracies achieved by a model for all the increments. For robustness, we ran all the experiments for all the models 10 times with different random seeds and report average and standard deviation of the accuracies. 


\begin{table*}
\centering
\small
\begin{tabular}{|P{1cm}|P{1.35cm}|P{1.3cm}|P{1.3cm}|P{1.3cm}|P{1.3cm}|P{1.3cm}|P{1.3cm}|P{1.3cm}|P{1.3cm}| }
     \hline
     \textbf{k-shot} & \textbf{Networks} & \textbf{FLB} & \textbf{FT} & \textbf{iCaRL} & \textbf{LWF} & \textbf{EWC} & \textbf{CWR} & \textbf{CBCL} & \textbf{EEC} \\
     \hline
     \multirow{3}{*}{5-Shot} & ResNet18 & 82.1$\pm$8.4 & 18.7$\pm$15.3 & 19.4$\pm$17.6 & 14.9$\pm$12.8 & 13.8$\pm$13.0 & 39.1$\pm$25.1 & 80.6$\pm$7.9 & 22.5$\pm$15.8\\
     & VGG16 & 83.8$\pm$7.2 & 23.5$\pm$21.2 & 24.4$\pm$25.9 & 14.8$\pm$16.4 & 16.9$\pm$16.9 & 46.3$\pm$22.8 & 78.1$\pm$9.2 & 24.4$\pm$21.0\\
     & ResNet50 & 85.1$\pm$7.5 & 17.6$\pm$15.8 & 34.5$\pm$16.5 & 14.1$\pm$15.1 & 17.2$\pm$22.2 & 39.2$\pm$24.5 & 85.3$\pm$5.9 & 17.3$\pm$15.8\\
     \hline
     \multirow{3}{*}{10-Shot} & ResNet18 & 89.2$\pm$5.2 & 22.1$\pm$19.6 & 19.0$\pm$18.0 & 18.7$\pm$19.6 & 15.4$\pm$18.6 & 39.5$\pm$24.6 & 86.6$\pm$5.9 & 22.8$\pm$17.7\\
     & VGG16 & 86.0$\pm$6.4 & 25.4$\pm$23.8 & 21.5$\pm$21.1 & 14.1$\pm$15.2 & 18.0$\pm$18.1 & 46.4$\pm$22.6 & 84.1$\pm$6.1 & 25.8$\pm$24.0\\
     & ResNet50 & 90.3$\pm$5.1 & 21.0$\pm$16.8 & 37.7$\pm$14.8 & 13.2$\pm$14.0 & 17.5$\pm$21.6 & 39.6$\pm$24.3 & 89.3$\pm$4.7 & 18.8$\pm$16.3\\
 \hline
 \end{tabular}
 \caption{Comparison of 8 incremental learning approaches on F-SIOL-310 using different backbone network architectures for 5-shot and 10-shot incremental learning with 2 classes per increment in terms of average incremental accuracy (\%). Mean and standard deviations of the average incremental accuracy in 10 runs are reported.}
 \label{tab:networks}
 \end{table*}

\subsection{Benchmarks for FSIL} 
\label{sec:few_shot_incremental}

Figure \ref{fig:few_shot_results} (a) compares the 8 SOTA incremental learning approaches with ImageNet features on F-SIOL-310 
for 5-shot (Figure \ref{fig:few_shot_results} (a) (left)) and 10-shot (Figure \ref{fig:few_shot_results} (a) (right)) incremental learning experiments in terms of classification accuracy. For both 5-shot and 10-shot incremental learning experiments, all the models except CBCL and FLB suffer from catastrophic forgetting. Even though iCaRL, EWC, LWF, EEC and CWR have been designed to reduce the effect of catastrophic forgetting (as demonstrated on simple large scale datasets like MNIST and CIFAR-100), they do not perform well on the FSIL scenario and perform similarly to FT which is the baseline for catastrophic forgetting. Note that iCaRL stores the complete data of all the previous classes, still it suffers from catastrophic forgetting, which demonstrates that iCaRL struggles with the  FSIL vision challenge. FLB, the batch learning baseline (it does not learn incrementally), uses the data of all the classes in each increment, hence it produces the best results. CBCL is the only method that produces favorable results on the two FSIL settings. Still, note that CBCL's accuracy drops on F-SIOL-310 compared to CIFAR-100 as reported in \cite{Ayub_2020_CVPR_Workshops}. In the original paper \cite{Ayub_2020_CVPR_Workshops}, CBCL outperforms FLB by a significant margin on FSIL experiments on CIFAR-100. However, on F-SIOL-310 it produces $\sim$9\% lower accuracy than FLB after learning all the classes. These results clearly show that the current incremental learning approaches are not suitable for few-shot incremental object learning for robotics applications. CBCL is the most promising method and it mitigates the effects of catastrophic forgetting, but it still produces only $\sim$70\% accuracy after learning only 22 object classes. Many realistic domestic robotics applications will likely demand a much higher object classification accuracy even after learning a large number of object classes.

\subsection{Effect of Using Different Base Features}
For the previous experiment all the SOTA approaches used ResNet-18 pre-trained on ImageNet as the feature extractor. To test the effect of the base features, we trained ResNet-18 from scratch on the CIFAR-100 dataset and use it as the backbone architecture for all the approaches. CIFAR-100 is a much smaller dataset than ImageNet containing only 50,000 training images belonging to 100 classes. In comparison, ImageNet contains 1.2 million images belonging to 1000 classes. Hence, the CIFAR-100 base features are not as general as ImageNet features.

Figure \ref{fig:few_shot_results} (b) compares the 8 incremental learning approaches on 5-shot and 10-shot incremental learning using ResNet-18 pre-trained on CIFAR-100. Similar to the previous experiment, all the models except CBCL and FLB suffer from catastrophic forgetting for both 5-shot and 10-shot incremental experiments. However, EEC and iCaRL show better performance ($\sim$10\%) than when using ImageNet features. Both of these approaches learn a representation and a classifier together for incremental learning. Hence, it is difficult for the representation to adjust after only a few training examples when starting from ImageNet features. However, for CIFAR-100 the features learned by the model are not as general (because of the smaller size of CIFAR-100) hence it is easier for the representation to adjust even with a few training examples.

Both FLB and CBCL show significantly lower accuracy ($\sim$12\%) when using CIFAR-100 features than with ImageNet features. Both approaches use the fixed features from the pre-trained network and do not learn their own representation. Since CIFAR-100 features are not as general as ImageNet features, both FLB and CBCL struggle in the FSIL scenario. These results further show the limitations of the current incremental learning techniques. Even methods that are specifically designed for FSIL, such as CBCL, depend on the availability of good base features.  

\subsection{Effect of Using Different Backbone Architectures}
Table \ref{tab:networks} compares the average incremental accuracies of 8 incremantal learning approaches on 5-shot and 10-shot incremental learning with three different backbone network architectures (ResNet18 \cite{He_2016_CVPR}, VGG16 \cite{Karen14}, ResNet50 \cite{He_2016_CVPR}) pre-trained on ImageNet dataset. Both FLB and CBCL produce the best results when using a deeper backbone like ResNet50, since both FLB and CBCL use the network as a fixed feature extractor and a deeper backbone has higher generalization power. iCaRL also gets the best results with ResNet50 utilizing the higher capacity of ResNet50 when learning using all the old and new class images. The other 5 approaches achieve slightly better results using VGG16 over other networks. However, all of the approaches (except FLB and CBCL) suffer from catastrophic forgetting regardless of the backbone network used. Note that the variance of average incremental accuracy for all the approaches (especially FT, iCaRL, LWF, EWC, CWR and EEC) is significantly high. This shows that all the approaches are highly susceptible to the order of the classes presented in the training set depicting that they are unreliable in real-world applications with unknown order of classes in different increments. Also, note that the average incremental accuracy for 5-shot and 10-shot learning is similar for all the approaches, except FLB and CBCL. These results show that a small increase in number of images per class from 5 to 10 does not improve the accuracy of these approaches since they all require a large number of training examples per class. These findings further show that using a deeper backbone network architecture can help some approaches achieve better performance. However, current SOTA approaches are still far from being deployed in real-world robotics applications. Hence, F-SIOL-310 can be used by future incremental learning approaches to close the gap between research and real-world robotics applications.

\section{CONCLUSION}
\label{sec:conclusion}
\noindent In this paper we have presented a new dataset (F-SIOL-310) specifically designed for the FSIL problem for robotics applications. The dataset encompasses many real-world challenges faced by robots, like different objects in training and test sets, scale variation and transparency in objects. Through extensive experiments on F-SIOL-310 using 8 incremental learning approaches, we show that the FSIL problem for robotics applications is far from being solved.  

Our results have shown that using deeper backbone architectures pre-trained on a large scale image dataset helps improve the performance of incremental learning techniques. Another finding from our results is that using a pre-trained network as a feature extractor produces better accuracy than finetuning the network. Among the incremental learning approaches, CBCL produces the best results because it was specifically designed for FSIL. Although CBCL produces comparable accuracy to the batch learning baseline when the number of classes is small ($\sim$10), its accuracy starts to drop below the batch learning baseline when the total number of classes learned increases ($\sim$22).








\section*{ACKNOWLEDGMENT}
\noindent This work was supported by Air Force Office of Scientific Research contract FA9550-17-1-0017.


{\small
\bibliographystyle{IEEEtran.bst}
\bibliography{main}
}

\end{document}